%% file: neurips_2021.tex
\documentclass{article}

\usepackage[nonatbib, preprint]{neurips_2021}

\usepackage[utf8]{inputenc} 
\usepackage[T1]{fontenc}    
\usepackage{hyperref}       
\usepackage{url}            
\usepackage{booktabs}       
\usepackage{nicefrac}       
\usepackage{microtype}      
\usepackage{xcolor}         
\usepackage{graphicx}
\usepackage{amsmath,amssymb,amsfonts,stmaryrd}
\usepackage{bm}
\usepackage{algorithmic}
\usepackage{textcomp}
\usepackage{multicol}
\usepackage{caption}
\usepackage{float}
\usepackage{cite}
\usepackage{subcaption}
\usepackage{wrapfig}
\usepackage[ruled,vlined]{algorithm2e}
\usepackage{glossaries}
\input{./acronyms.tex}

\title{Tackling Asymmetric and Circular Sequential Social Dilemmas with Reinforcement Learning and Graph-based Tit-for-Tat}

\author{%
  Tangui Le Gléau\\
  Orange Labs Lannion\\
  \texttt{tangui.legleau@gmail.com} \\
  \And
  Xavier Marjou\\
  Orange Labs Lannion\\
  \texttt{xavier.marjou@orange.com} \\
  \And
  Tayeb Lemlouma\\
  IRISA\\
  \texttt{tayeb.lemlouma@irisa.fr} \\
  \And
  Benoit Radier\\
  Orange Labs Lannion\\
  \texttt{benoit.radier@orange.com} \\
}
\begin{document}

\maketitle

\begin{abstract}

In many societal and industrial interactions, participants generally prefer their pure self-interest at the expense of the global welfare. Known as social dilemmas, this category of non-cooperative games offers situations where multiple actors should all cooperate to achieve the best outcome but greed and fear lead to a worst self-interested issue. Recently, the emergence of Deep \gls{rl} has generated revived interest in social dilemmas with the introduction of \gls{ssd}. Cooperative agents mixing RL policies and Tit-for-tat (TFT) strategies have successfully addressed some non-optimal Nash equilibrium issues. However, this kind of paradigm requires symmetrical and direct cooperation between actors, conditions that are not met when mutual cooperation become asymmetric and is possible only with at least a third actor in a circular way. To tackle this issue, this paper extends \gls{ssd} with \gls{cssd}, a new kind of Markov games that better generalizes the diversity of cooperation between agents. Secondly, to address such circular and asymmetric cooperation, we propose a candidate solution based on \gls{rl} policies and a graph-based TFT. We conducted some experiments on a simple multi-player grid world which offers adaptable cooperation structures. Our work confirmed that our graph-based approach is beneficial to address circular situations by encouraging self-interested agents to reach mutual cooperation.

\end{abstract}

\section{Introduction}
In many everyday situations involving several participants, like environmental issues, self-interest leads to a selfish behavior while global cooperation would allow a better outcome. Such behavior, driven by the temptation to exploit cooperators or by the fear not to be followed by others, leads to defective behavior. Those situations, known as social dilemmas, exist as soon as multiple non-cooperative intelligent agents interact with each other. They happen in various multi-agent systems: competitive actors exchanging resources (electricity or connectivity providers, data for learning devices, etc.) or robotics (multiple independent autonomous vehicles sharing the same environment). As a consequence, the study of incentives is increasingly strategic in view of the proliferation of non-cooperative multi-actor situations and growing socio-environmental challenges \cite{hager2019artificial, dafoe2020open}.

In recent decades, the theory of social dilemmas has attracted much interest. From the notion of general-sum games \cite{nash1951non} and the formulation of prisoner's dilemma \cite{flood1958some} to the iterated prisoner's dilemma tournaments of Axelrod \cite{axelrod1981evolution}, literature has been developed in the field. A major part of the field of social dilemmas deals with matrix games with atomic actions (cooperation/defection) with the introduction of three canonical social dilemmas (the stag hunt, the chicken game and at last the so-called prisoner's dilemma). 

Numerous methods have been proposed to solve such matrix games. The \gls{tft} \cite{rapoport1965prisoner} known as the winner of the Axelrod's tournament \cite{axelrod1981evolution} is rather simple: begin with cooperation and then reproduce the previous choice of the partner. Since then, some variants of \gls{tft} \cite{verhoeff1998trader, beaufils2001adaptive} of alternatives such as win–stay, lose–shift strategies \cite{nowak1993strategy} have also been proposed. At last, \gls{rl} methods have been studied in the iterated prisoner's dilemma \cite{sandholm1996multiagent, izquierdo2008reinforcement, de2006learning}, with a conclusion that \gls{rl} agents struggle to avoid Nash mutual defection.

Then recently, a second major part of interest has emerged: with the development of Deep Reinforcement Learning policies \cite{mnih2013playing, mnih2015human}, a new kind of games was proposed: \cite{leibo2017multi} introduced sequential social dilemmas, a version of social dilemma where Deep \gls{rl} policies replace atomic actions (cooperate/defect) allowing more realistic scenarios. While Deep RL was originally known to tackle zero-sum games \cite{silver2016mastering}, the coordination of Deep RL agents has also been studied in general-sum games first in situations where the cooperation was encouraged \cite{tampuu2017multiagent, foerster2016learning, lowe2017multi, peysakhovich2017prosocial} and then also in general-sum games with no incentive to cooperate (the so-called social dilemmas). At last, to address the issue of non-optimal Nash equilibrium and avoid the mutual defective behavior, \cite{lerer2017maintaining} proposed a safe and incentive approach mixing \gls{rl} policies and a TFT strategy.

\begin{wrapfigure}[21]{r}{0.42\columnwidth}
    \includegraphics[trim=0cm 10cm 27cm 0.6cm, width=0.285\columnwidth]{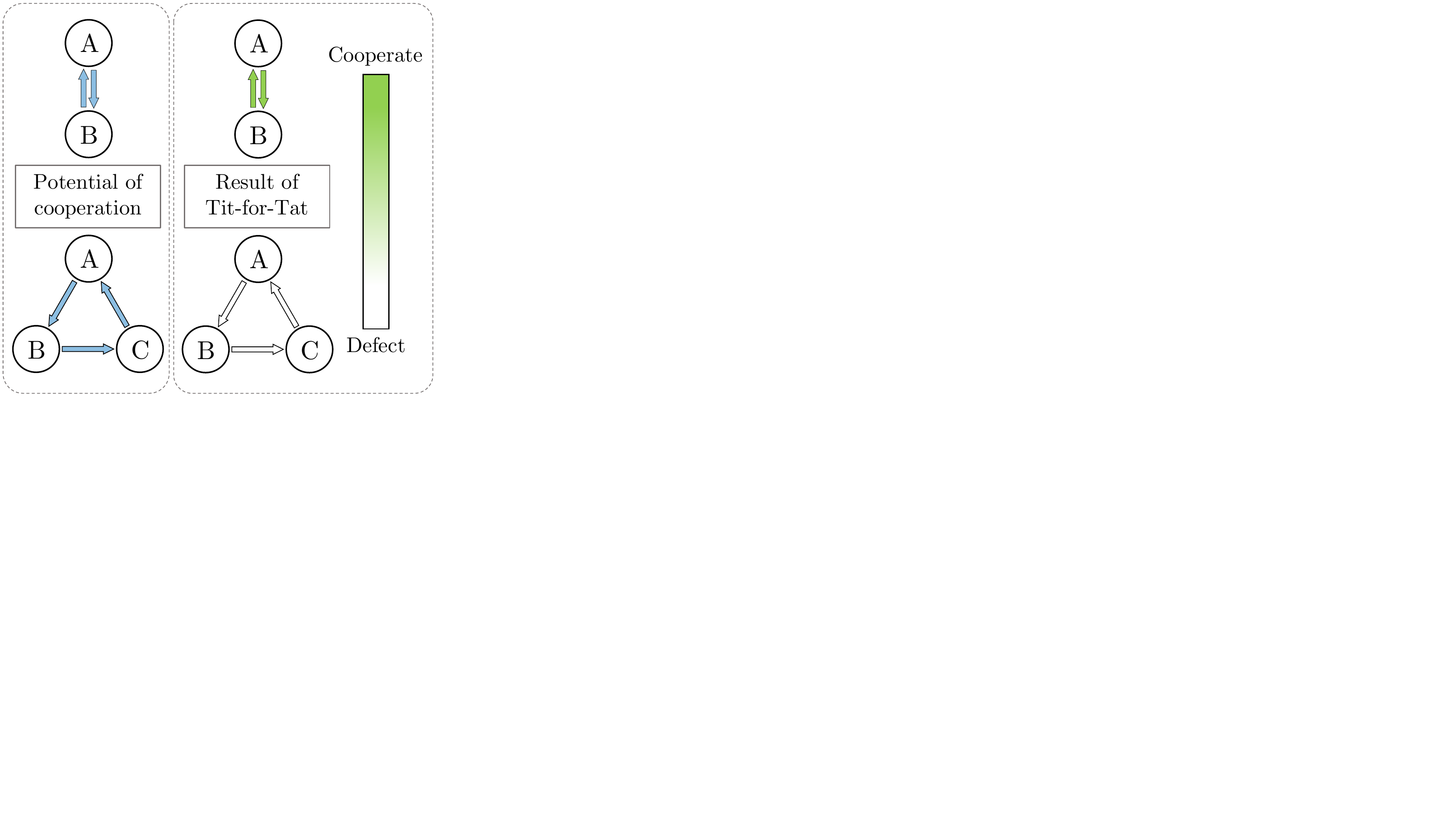}
 \captionsetup{justification=justified}
    \caption{When the potential cooperation is bilateral, two partners can make a cooperation emerge by TFT; however when the potential cooperation is possible only in a circular way, vanilla TFT can't converge since it has not direct response.}
      \label{fig:cir_TFT_pb}
\end{wrapfigure}

The previous approaches uses vanilla \gls{tft}, which requires direct bilateral cooperation between two agents. However, there are numerous situations where vanilla \gls{tft} does not work as agents can not give back what they received to their partner agent. Yet, such non-bilateral situations often leave room for indirect cooperation between agents so that cooperation is achieved in a circular way. Therefore, to study this particularity, we introduce the \gls{cssd}, a kind of Markov game which is an extension of the sequential social dilemma of \cite{leibo2017multi} where a potential of cooperation between agents can be represented by a weighted directed graph in which it can exist some cycles of cooperation. Such a model is adapted to better generalize possible cooperation and emphasize the distinction of players in games, in particular detecting precise free riders, which is impossible in a one-vs-one or one-vs-all-others approach. We bring a formalism for this kind of problem and we introduce in the paper one adaptable circular Markov game we designed for the study of this new paradigm.

To tackle such games, we propose an algorithm that extends the approach of \cite{lerer2017maintaining}. Based on a mix between \gls{rl} policies and a graph-based approach of \gls{tft}, we introduce an agent whose goal is to find an optimal cooperation between more than two agents, while preventing being exploited by defectors and, if possible, maintain a cooperation without defectors when a cooperation cycle exist. 

We conducted some experiments on different cooperation structures and observed the results with three designed social metrics (efficiency, incentive-compatibility and safety). Our first conclusions are that the graph structure of our approach is relevant since the original approach is inefficient in non-bilateral situations. Even if our approach struggles to detect optimally the cooperation graph of agents, it manages to performs better that egoist behavior. We are convinced that the model offers a lot of perspectives and possibilities of improvements.




\section{Circular sequential social dilemmas} 
\label{sec:social_dilemma}
In this section, we recall the model of Markov game and detail the basic definitions of a basic social dilemma, and then introduce our definition of the circular sequential social dilemma.

\subsection{$N$-player Markov games}
\label{sec:markov_games}
To model our games, we use stochastic Markov games, which are $N$-player Partially Observable Markov Decisions Processes (POMDP) \cite{shapley1953stochastic, littman1994markov}. A $N$-player Markov game $\mathcal{M}$ is defined by an uplet $(\mathcal{I}, \mathcal{S}, \mathcal{A},O,\mathcal{T},R)$ where $\mathcal{I} = \{1,..., N\}$ is a set of players, $\mathcal{S}$ is the set of states and $O : \mathcal{S} \times \mathcal{I} \rightarrow \mathcal{S}$ is an observation function. $\mathcal{A} = \mathcal{A}^1 \times ... \times \mathcal{A}^N$ is the set of joint actions, and a joint action $\vec{a} = (a^1, ..., a^N)$ makes the state $s$ of the game change following the stochastic function $\mathcal{T} : \mathcal{S} \times \mathcal{A}^1 \times ... \times \mathcal{A}^N \rightarrow \Delta(\mathcal{S})$. At last, a personal reward function $R : \mathcal{I} \times \mathcal{S} \times \mathcal{A} \rightarrow \mathbb{R} $ gives a reward $r^i$ to each player $i$ (the joint reward is denoted $\vec{r} = (r^1,...,r^N)$). Each agent $i$'s goal is to find a policy $\pi^i : \mathcal{S} \rightarrow \Delta(\mathcal{A}^i)$ ($\vec{\pi}$ denotes the joint policy $(\pi^1,...,\pi^N)$) in order to maximize the expected $\gamma$-discounted reward defined by:
\[ V^i_{\vec{\pi}}(s_0) = \mathbb{E}[\sum_{t=0}^{\infty} \gamma^t r^i(s_t,\vec{a}_t) | \vec{a}_t \sim \vec{\pi}, s_{t+1} \sim \mathcal{T}(s_t, \vec{a}_t)   ]   \]

\subsection{Social dilemmas}
\label{sec:social_dilemma}

Repeated $2$-player matrix games provide a basic framework for social dilemmas: each player chooses between two possible actions (cooperate or defect). Table \ref{tab:SD_payoffs} summarizes the four possible payoffs which are generally denoted by R (Reward for mutual cooperation), S (Sucker outcome of a cooperator exploited by a defector), P (Punishment for mutual defection) and T (Temptation to exploit a cooperator).

\begin{table}[h!]
  \caption{Payoffs in a 2-player social dilemma}
  \label{tab:SD_payoffs}
  \centering
  \begin{tabular}{lcc}

    \toprule
          & Cooperate     & Defect \\
    \midrule
    Cooperate & $(R,R)$  & $(S,T)$     \\
    Defect     & $(T,S)$ & $(P,P)$     \\
    \bottomrule
  \end{tabular}
\end{table}

This matrix game is defined as a social dilemma if the four payoffs of Table \ref{tab:SD_payoffs} verify the following inequalities \cite{macy2002learning}:

\begin{enumerate}
    \item $R > P$: Mutual cooperation better than mutual defection
    \item $R > S$: Mutual cooperation better than exploitation 
    \item At least one of these two inequalities:
    \begin{itemize}
        \item $T>R$: greed (3a)
        \item $P>S$: fear (3b)
    \end{itemize}{}
    \item $R>\frac{1}{2}(S+T)$: mutual cooperation is better than equiprobable different choice (4)

\end{enumerate}{}

A social dilemma admits at least one non-optimal Nash equilibrium in particular (\textit{Defect, Defect}) in the Prisoner's Dilemma (where greed (3a) and fear (3b) are verified).

\subsubsection{Continuous cooperative policies}
\label{sec:cont_social_dilemma}
Before defining our novel model in section \ref{sec:circular_SSD}, we need to introduce our formalism of policies of graded cooperation which is a slight extension of the cooperative policies proposed by \cite{leibo2017multi}. Let us assume that each player is free to choose a degree of cooperation towards each of the other players. Formally, if we denote the set of policies $\Pi^{(i)} = \Delta(\mathcal{A}^i)^S$ for the player $i$, then we empirically define the policy $\pi^i_{\overrightarrow{C_i}} \in \Pi^{(i)}$ as the policy for player $i$ in which he has a graded cooperative behavior towards $j$ described by the degree $c_{i,j}$ given by the vector $\overrightarrow{C_i} = (c_{i,j})_j$ (between 0 for full defection and 1 for full cooperation). Moreover, the value function for each player $i$ corresponding to cooperation degrees is denoted $G^{(i)}: [0,1]^{N\times N}\times \mathcal{S} \rightarrow \mathbb{R}$. It means that at state $s$, player $i$ earns an expected payoff defined by the value function (\ref{sec:markov_games}): $G^{(i)}((c_{i,j})_{i,j},s) = V^i_{\vec{\pi}}(s)$ where $\vec{\pi} = (\pi^j_{\overrightarrow{C_j}})_j$ depends on the $N^2$ cooperation degrees (by convention $c_{i,i}=1$).

\newpage

\subsection{Circular sequential social dilemmas}
\label{sec:circular_SSD}

We introduce our novel notion of circular sequential social dilemma. With the formulation of \ref{sec:cont_social_dilemma}, let the tuple $(\mathcal{M},\Pi, G)$ be a circular sequential social dilemma if there exist states $s\in \mathcal{S}$ for which it exists at least one finite sequence ${(i_{k})}_{k \in \llbracket 0, K-1 \rrbracket} $ indexing $2< K\leq N$ distinct players of $\mathcal{M}$ such as: \newline \newline 
For all $k \in \llbracket 0, K-1 \rrbracket$ (by abuse of notation: $ k \equiv i_k $ for better readability and with $i_K = i_0$ for circularity):
\begin{enumerate}
    \item $G^{(k)}$ is constant w.r.t. $c_{k+1,k}$ \\ i.e player $k+1$ can't directly help $k$ whatever its willingness to cooperate
    \item $G^{(k)}$ is increasing w.r.t. $c_{k-1,k}$ \\ (temptation to exploit or greed)
    \item $G^{(k)}$ is decreasing w.r.t. $c_{k,k+1}$ \\ (fear of being exploited)
    \item It exists values $\mathcal{C} \in [0,1]^{N \times N} $ such as $\frac{\partial G^{(k)}}{\partial c_{k,k+1} }(\mathcal{C}) + \frac{\partial G^{(k+1)}}{\partial c_{k,k+1}}(\mathcal{C}) > 0$\\  (positive-sum property: the cost of the "giver"  is lower than the gain of the "receiver")
\end{enumerate}
Let us notice that the above conditions may also characterize the classic two-player symmetrical \gls{ssd} ($K=2$) if the first condition is removed.

\begin{wrapfigure}[28]{r}{0.58\columnwidth}


        \includegraphics[trim=0.2cm 4cm 15.3cm 0.15cm, clip,width=0.58\columnwidth]{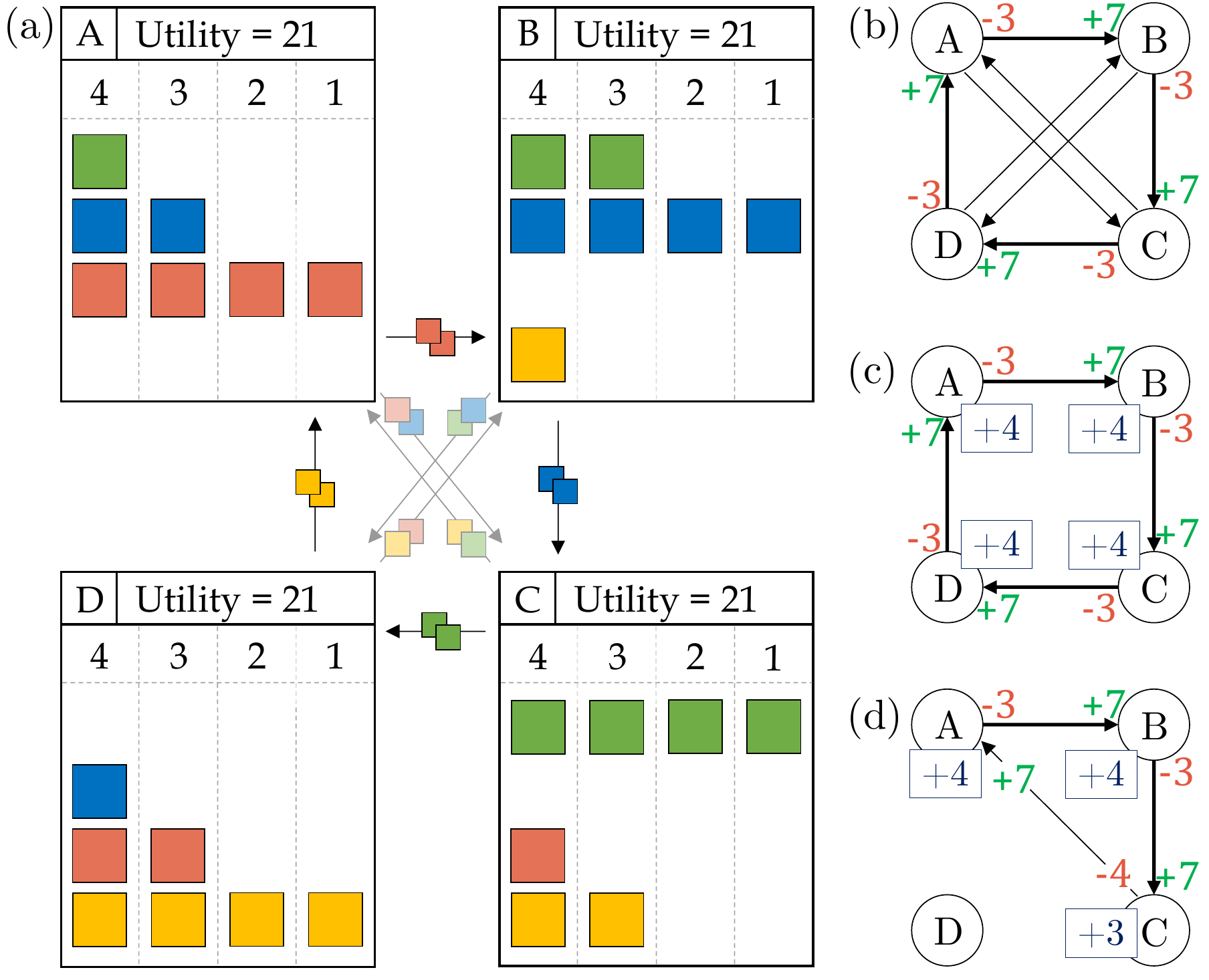}

 \caption{Sharing resources whose marginal utility is decreasing (e.g. $5-k$ for the $k^{th}$ item). (a) designs the current quantity of items for the agents, showing the potential optimal 4-agent cycle of cooperation. Graph (b) shows the potential gain and cooperation, (c) and (d) are respectively situations with four and three cooperators: in red the cost of the giver and in green the gain of the receiver; boxed values indicate the gain of social welfare.}
  \label{fig:explications_sharing}

\end{wrapfigure}
 
 For better understanding and intuition, let us focus on a situation of resource sharing: multiple agents have items (e.g. perishable resources like electricity, energy, connectivity etc.) whose marginal utility is decreasing. In Figure \ref{fig:explications_sharing}, four agents A, B, C and D own items of different color whose marginal utility is decreasing (4 for the first item, 3 for the second...). In some situations, it is interesting for agents to reallocate their resources so that they all increase their outcome. However, if we assume that agents are self-interested, the situation becomes a dilemma since it is unsafe to cooperate alone and tempting to exploit cooperators. Moreover, in some situations, the dilemma can not be managed bilaterally and becomes circular. For example, in Figure \ref{fig:explications_sharing}, it exists some cycles of cooperation, e.g. in the cycle $<A,B,C,D>$, each player can help the next player (in a sense that there is a gain of the sum of utility), but can't directly help the previous one. Note that in case of a defector, other less optimal 3-player cycles are possible. 

\section{Related work}
\label{sec:related_work}
Social dilemmas have been studied for many years. First, the \gls{pd} was introduced by \cite{flood1958some} and a tournament of iterated Prisoner's Dilemmas was proposed by Axelrod \cite{axelrod1981evolution} in which the winner was the \gls{tft} \cite{rapoport1965prisoner}. Then, extensions of the \gls{ipd} have been proposed, in particular a continuous version of \gls{pd} as well as corresponding continuous versions of \gls{tft} \cite{verhoeff1998trader}. Recently, with the emergence of Deep \gls{rl}, \cite{leibo2017multi} has introduced sequential social dilemmas and has presented a study of \gls{marl}. \cite{hughes2018inequity} extended the formal definition of SSD to a $N$-player sequential social dilemma, with an one-vs-all-others approach, which differs slightly from our formalism that allows the distinction of players.
\cite{lerer2017maintaining} showed empirically that \gls{tft} was necessary to ensure the safety of policies, it also presented an approach mixing \gls{tft} and \gls{rl} policies. Also convinced of the necessity of \gls{tft} since pure \gls{rl} policies converge to non-optimal Nash Equilibrium, our approach is also inspired from such hybrid method. However, instead of a one-vs-one or one-vs-all-others point of view, our approach differs by the distinction of the different players. In addition, our proposed approach also involves a circular version of \gls{tft}.
Other works have tackled the issue of cooperation between \gls{rl} agents: an intrinsic motivation in the reward to encourage cooperative behaviors \cite{jaques2019social}, a modified Actor-Critic to improve coordination of multiple agents in cooperative and competitive scenarios \cite{lowe2017multi} or the study of the \gls{marl} paradigm in situation of common resource appropriation \cite{perolat2017multi}. Finally, \cite{li2020end} proposed a completely different approach which consists in learning to intervene to incentive agents. However, this approach needs a central third party to modify environments and payoffs which doesn't fit with our objective.

\section{Our approach}
\label{sec:approach}
We introduce our approach, the \textsc{grTFTrl}, which extends the algorithm of \cite{lerer2017maintaining} with more than two agents and with the possibility to address asymmetric and circular cooperation. It relies on \gls{rl} policies of different cooperation degrees which are trained in an offline selfplay phase and a graph-based \gls{tft} that selects the best cooperative policy. We detail our propositions for the different components of the method that we gather in the algorithm \ref{algo:grTFTrl} at the end of this section\footnote{Our implementation is available here: \url{https://github.com/submission-conf/neurips_cooperativeAI}}.

\subsection{\gls{rl} background}
\label{sec:bg_rl}
\gls{rl} is a way to train policies $\pi : s_t \rightarrow a$ which are functions that map an observation $s_t$ to an action $a$ in order to maximize a total payoff $V_{\vec{\pi}}(s_0)$ \cite{sutton1998introduction} (see section \ref{sec:markov_games}). Numerous techniques exist, however the choice of the training algorithm is out of scope of the paper. We only assume that our policies are associated to Q values which are functions that map states and action to the estimated payoff: $Q[s_t, a]$. Therefore, the training algorithm can be a vanilla Deep Q-network \cite{mnih2013playing} or an Actor-Critic method \cite{mnih2016asynchronous} whereas simple Policy Gradient methods can not be used.

\subsection{\gls{tft} background}
\label{sec:bg_tft}

The principle of \gls{tft} is to react to a cooperation choice from a partner in order to incentive cooperation and be robust to defection. The discrete principle is rather simple \cite{rapoport1965prisoner}: begin with cooperation and reproduce the choice of the partner. Several approaches have been proposed \cite{verhoeff1998trader} for continuous versions. We propose to gather several existing contributions in this formulation:
\begin{equation}
\begin{aligned}
&\text{TFT}_{\alpha, \beta, \gamma, r_0}(t, a_{t-1}, b_{t-1}), r_t = 
     \begin{cases}
       0,r_0 &\quad\text{if } t = 0 \\
       \alpha a_{t-1} + (1-\alpha) (r_{t} + (1-r_{t})b_{t-1}),\\
       [r_{t-1} + \beta(b_{t-1} - a_{t-1})]^+ + r_0 \mathcal{B}(1,\gamma)
       &\quad\text{if } t > 0.
     \end{cases}
\end{aligned}
\end{equation}
In the formulation, $\text{TFT}(t, a_{t-1}, b_{t-1})$ allows to compute an ideal cooperation degree $a_t$ that a player A should choose for a safe and incentive response to partner B according to the previous step (degrees $a_{t-1}$ and $b_{t-1}$). It has an inertia coefficient $\alpha$ and uses to encourage cooperation a dynamic incentive $r_t \in [0,1]$ that is modified with an adaptive coefficient $\beta$ and a Bernoulli variable of parameter $\gamma$.



\subsection{Policies generation by \gls{rl} selfplay}
One major part of our approach relies on an offline training phase in which some policies with different cooperation degrees are trained. We assume that agents are able to generate a policy $\pi^i_{\overrightarrow{C}} $ defined by a cooperation vector $\overrightarrow{C} \in [0,1]^{N}$ describing how an agent $i$ should interact with each of the other agents. The policies $\pi^i_{\overrightarrow{C}} $ are trained by \gls{rl} methods with a reward that is modified in accordance with the cooperation objective: $\Tilde{r}^i = \langle \overrightarrow{C},\overrightarrow{r}\rangle = \sum_{j=1}^N  \overrightarrow{C}[j]~r^j$ where $\overrightarrow{r}$ is the joint reward earned by the $N$ agents (defined in section \ref{sec:markov_games}). 
In theory, we use the formalism $\pi^j_{\overrightarrow{C}} $, but in practice, agents only train a set of policies of discrete cooperation vectors: $\Pi^{(i)} = \{ \pi^i_{\overrightarrow{C}}, \overrightarrow{C} \in \{0,1\}^{N} \}$.


\subsection{Detection of cooperation}
\label{sec:detection_coop}

In our approach, agents need to detect the maximal potential cooperation graph, as well as the current cooperation between agents. Therefore we assume that our agent updates at each step, an inner graph of maximal potential cooperation $\mathcal{G}_{max}^{t}$ and an inner graph of the current cooperation $\mathcal{C}^{t}$. First, let us recall that each agent can observe the full state $s_t$ and the previous actions $\overrightarrow{a_{t-1}}$. Moreover, thanks to \gls{rl} selfplay, agents have access to some policies. In particular, let be $Q_{k \rightarrow l}$ the Q values of the policy of agent $k$ which cooperates with $l$ (i.e. $Q^k_{\overrightarrow{C}}$ where $\overrightarrow{C}$ has values 1 at indices $\{k,l\}$ and 0 elsewhere), as well as $\pi_{k \rightarrow l}$ the associated policy and $V_{k \rightarrow l}[s_t] = \max\limits_{a} Q_{k \rightarrow l}[s_t, a]$ its value function. 

To detect the maximum potential cooperation graph $\mathcal{G}_{max}$, we propose to apply a soft-update with the normalized difference of social welfare between the cases of cooperation $i \rightarrow i$ and $i \rightarrow j$:

\begin{equation}
\label{eq:g_max}
\begin{aligned}
&\mathcal{G}_{max}^{t+1} = (1-\tau)\mathcal{G}_{max}^{t} + \tau\mathcal{G}, \quad \text{with} ~ \mathcal{G}[i,j] = \left[  \frac{  V_{i \rightarrow j}[s_t] -   V_{i \rightarrow i}[s_t]   }{   \max\limits_{k} V_{i \rightarrow k}[s_t]   -   V_{i \rightarrow i}[s_t]   } \right]^+
\end{aligned}
\end{equation}


As for the current cooperation, we propose to define the detected cooperation of $i$ towards $j$ as the positive part of normalized difference between the expected return received by agent $i$ following a cooperative policy towards $j$ and that of self-interested policy. Then, a soft-update is applied:

\begin{equation}
\label{eq:C_det}
\begin{aligned}
& \mathcal{C}^{t+1} = (1-\tau)\mathcal{C}^{t} + \tau C^{t-1}, ~\quad \text{with} ~ C^{t-1}[i,j] = \left[ \frac{Q_{i \rightarrow j}[s_{t-1}, a^i_{t-1}] - Q_{i \rightarrow i}[s_{t-1}, a^i_{t-1}] }{V_{i \rightarrow j}[s_{t-1}]-V_{i \rightarrow i}[s_{t-1}]} \right]^+
\end{aligned}
\end{equation}


\subsection{Graph-based TFT}
\label{sec:grTFT}

According to a maximal potential cooperation graph $\mathcal{G}_{max}$ and the previous cooperation degrees of agents $\mathcal{C}^{t-1}$, agents need to make an optimal choice of cooperation degrees towards other agents following a \gls{tft} approach (optimal, safe and incentive). Therefore, for this purpose we use a graph-based \gls{tft}, a work from previous work (under review for which we provide details in Supplementary Material). We simply here define the grTFT as a function with as input a maximal cooperation graph $\mathcal{G}_{max}$ and the previous cooperation graph $\mathcal{C}^{t-1}$ and returns the vector $\overrightarrow{C}$ of ideal cooperation degrees:
\begin{equation}
\begin{aligned}
&\text{grTFT} : \mathbb{N}\times \{1,N\} \times [0,1]^{N\times N} \times [0,1]^{N\times N} \rightarrow [0,1]^{N} \\
&\text{grTFT}(t, i, \mathcal{G}_{max}, {\mathcal{C}}^{t-1}) = \overrightarrow{C}
\end{aligned}
\end{equation}
 \newline
This function uses a graph theory flow network approach and needs as component a vanilla \gls{tft} algorithm that can be chosen in the section \ref{sec:bg_tft}, and whose parameters will be evaluated in \ref{sec:tft_parameters}.

\newcommand\colorCO{gray}
\SetKwComment{Comment}{$\color{\colorCO}\triangleright$\ }{} 

\begin{algorithm}[H]
\SetAlgoLined
\textbf{Input:} $\forall j$, a set of discrete policies $\Pi^{(j)} = \{ \pi^j_{\overrightarrow{C}} \} $ and the associated $Q^{j}_{\overrightarrow{C}} $\\
Initialize $\mathcal{G}_{max} \leftarrow I_N$, $\mathcal{C}^{0} \leftarrow I_N$\\
\For{$t \in [1, T_{max}]$}{
  Update the maximal potential of cooperation $\mathcal{G}_{max}$\\
  Detect and update the cooperation graph ${\mathcal{C}}^{t-1}$\\
  Apply the graph-based TFT: $\overrightarrow{C} =$ grTFT$(t, i, \mathcal{G}_{max}, {\mathcal{C}}^{t-1}  ))$\\
  \If { $t \equiv 0 \pmod{K}$ }{ 
           Modify choice of partners : $\overrightarrow{C_{dis}}[k] = \mathcal{B}(1, {\overrightarrow{C}[k]} )$
        }
  Get the policy $\pi^i_{\overrightarrow{C_{dis}}}$ following the (discrete) degrees ${\overrightarrow{C_{dis}}}$\\
  Choose action $a^i \leftarrow \pi^i_{\overrightarrow{C_{dis}}}(s_t)$
}
\caption{Algorithm \textsc{grTFTrl} for agent $i$}
\label{algo:grTFTrl}
\end{algorithm}

\newpage
\section{Experiments and methods}
\label{sec:experiments}
In this section, we describe the game \textsc{Collect} we designed so that it can be conferred any structure of maximal potential cooperation, in particular circular cooperation\footnote{Circular games available here: \url{https://github.com/submission-conf/circular_games} }. We also detail the metrics used to evaluate the parameters of our approach.

\subsection{The Collect game}
The \textsc{Collect} game is a grid-world game where four agents can move in their own room. Coins of different color appear and disappear stochastically. The agents can collect coins of any color, if one coin of their color is collected (by themselves or by another agent), they receive a reward of $+2$, and if they collect one coin of different color, they get a reward of $-1$. Therefore, the dilemma relies on the fact that the cost of a helper is lower than the gain earned by the partner. 

\begin{wrapfigure}[26]{r}{0.5\columnwidth}


        \includegraphics[trim=0.0cm 1.0cm 29.0cm 0.1cm, clip,width=0.5\columnwidth]{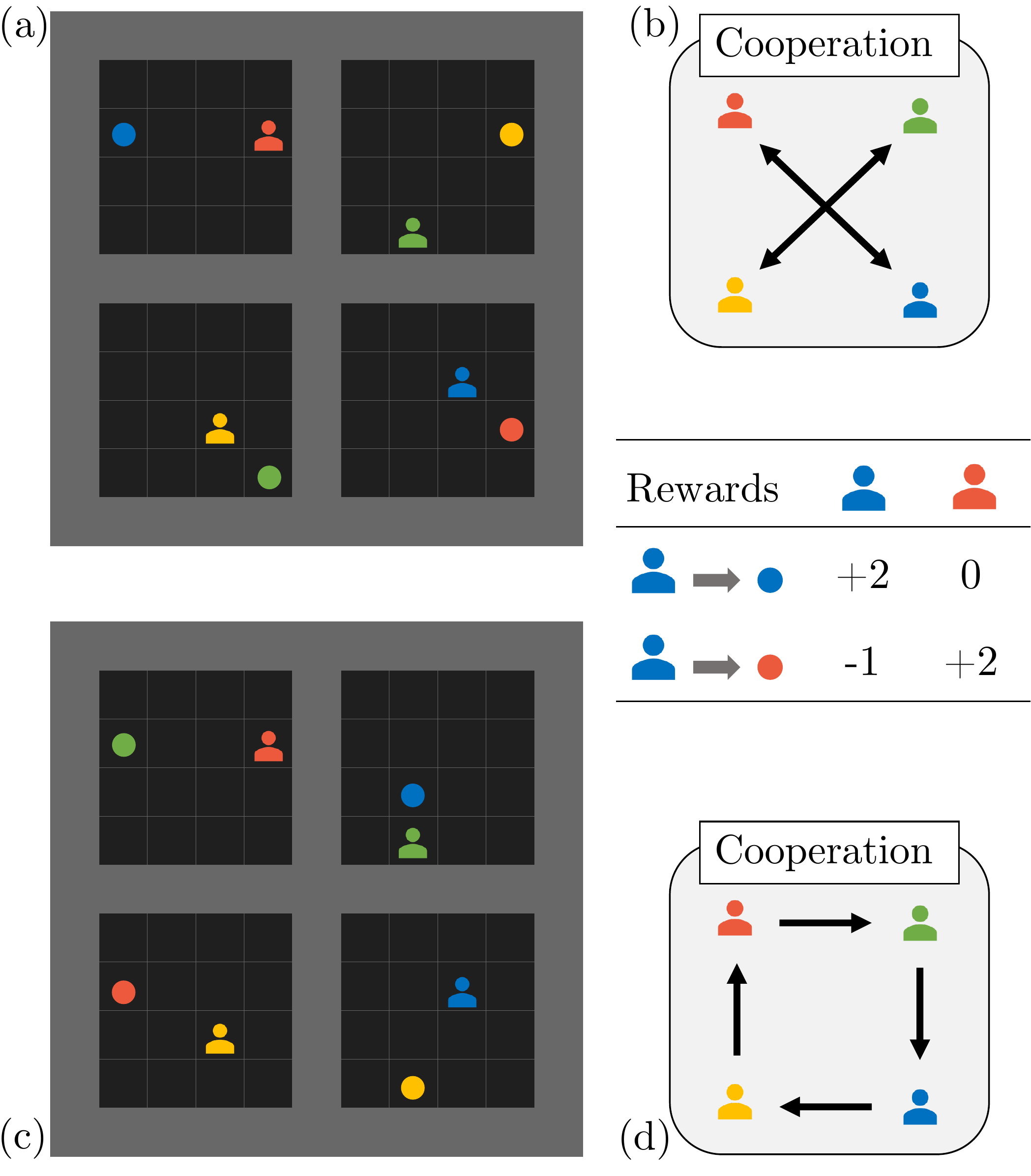}

 \caption{Two examples of the \textsc{Collect} game: the bilateral configuration (a) with the associated cooperation graph (b) where coins appear in a way that there are two pairs of bilateral cooperation. And a circular situation (c) with its circular cooperation graph (d).}
    \label{fig:example_collect}

\end{wrapfigure}

The size of the grid and rooms as well as the number of coins per rooms can be modified. However, the number of agents is fixed to four for simplicity and because it enables numerous interesting situations. We assume that agents can fully observe the full grid and all actions of agents. The disappearance of coins is governed by an exponential distribution and when a coin disappears or is collected, another coin reappears so that there is exactly a fixed number of coins in each room. The probabilities of the selection of color are given by a stochastic matrix explained in Section \ref{sec:collect_graph}.

\subsubsection{Configurable cooperation structure}
\label{sec:collect_graph}
What is interesting in the game is that the maximal cooperation graph is given by the stochastic matrix $P$ whose coefficients $P(i,k)$  define the probability that a coin appearing in the room of agent $i$ is of color $k$. This game parameter $P$ can be then configured to confer to the game any structure of potential cooperation between agents. For example, Figure \ref{fig:example_collect} presents two possible cases: a bilateral structure which can view as two 2-player social dilemmas and another configuration which is a pure circular dilemma. At last, $P$ can be dynamically modified during the game, to switch the cooperation structure.
\subsection{Social metrics}
We assume that $G^i(X_1, ..., X_N)$ refers to the sum of rewards over time of an agent $i$ when each agent $j$ uses policy $X_j$ in a game of $T$ steps, and that $C$ and $D$ denote the policies of naive cooperation and egoist defection. The \textit{Utilitarian metric} measures the average total welfare of agents, the \textit{safety} $Sf$ measures how it is interesting to choose one policy $\pi$ instead of defection while facing defectors. At last, the \textit{Incentive-compatibility} $IC$ measures how policies $\pi$ encourage an agent to choose cooperation instead of defection.
\[
    U(\pi) = \mathbb{E}\left[ \frac{\sum_{i=1}^N G^i(\pi, ..., \pi)}{T} \right], \quad Sf(\pi) = \mathbb{E}\left[ \frac{ G^1(\pi, D, ..., D) - G^1(D, D,..., D)}{T} \right],
\]

\[
 IC(\pi) = \mathbb{E}\left[ \frac{ G^1(C, \pi,..., \pi) - G^1(D, \pi,..., \pi)}{T} \right]
\]

\section{Results and discussion}
\label{sec:results_discussion}
Our approach is made up of several components which can each be evaluated. We present here some evaluations of the impact of choice of the \gls{tft} algorithm chosen to negotiate as well as the functions of detection of cooperation (the graph of cooperation between agents as well as the maximal potential graph imposed by the game).

\subsection{Graph-based \gls{tft} performances}
We evaluated our algorithm \textsc{grTFTrl}  and compared it with a baseline \textsc{Egoist}  which is a simple \gls{rl} policy (DQN) without any TFT algorithm. We also compare it with  \textsc{TFTrl} which is a similar approach but with a vanilla \gls{tft} without graph structure. Finally, we add in the evaluation a  \textsc{Nice} algorithm (an instance of \textsc{grTFTrl} with a naive cooperative grTFT) to emphasize the non-optimal safety and incentive-compatibility of an unconditional cooperation\footnote{Videos of some simulations are available here:\newline \url{https://youtube.com/playlist?list=PLzmQsQrITrGI16_-nj2qSzgTMHN1hCIoq} }.

\begin{table}[h!]
\small
  \caption{Results with environments \textsc{Bilateral} and \textsc{Circular}} (metrics $\times 100$, $5$ runs of $500$ steps )
  \label{tab:results1}
  \centering
  
  \begin{tabular}{lcccccc}\toprule
& \multicolumn{3}{c}{\textsc{Bilateral}} & \multicolumn{3}{c}{\textsc{Circular}} 
\\\cmidrule(lr){2-4}\cmidrule(lr){5-7}
           & $U$  & $IC$ & $Sf$    & $U$  & $IC$ & $Sf$ \\\midrule
 \textsc{Egoist}   & $2.5\pm 0.5$ & $-15.7\pm 2.4$ & $\bm{0.3\pm 0.7}$ & $2.1\pm 0.5$ & $-15.5\pm 1.8$ & $-1.0\pm 0.5$  \\
 \textsc{TFTrl} & $64.1\pm 5.2$ & $12.5\pm 2.5$ & $-1.6\pm 1.3$ & $5.8\pm 0.9$ & $-18.0\pm 2.9$ & $-1.2\pm 1.0$ \\
 \textsc{grTFTrl} & $67.1\pm 8.0$ & $\bm{16.8\pm 3.1}$ & $-1.2\pm 1.4$ & $65.3\pm 1.4$ & $\bm{18.8\pm 4.1}$ & $\bm{-0.4\pm 0.5}$  \\
 \textsc{Nice}   & $\bm{67.8\pm 2.4}$ & $-17.8\pm 8.4$ & $-15.2\pm 0.5$ & $\bm{67.8\pm 2.4}$ & $-17.8\pm 8.4$ & $-15.2\pm 0.5$ \\\bottomrule
\end{tabular}
  
\end{table}

The first observation is that the addition of a \gls{tft} provides a safer and more incentive agent. However, in case of circular cooperation,  \textsc{TFTrl} with vanilla \gls{tft} fails to cooperate whereas our approach can pursue safe cooperation.

\subsection{Impact of \gls{tft} parameters }
\label{sec:tft_parameters}
As mentioned in Section \ref{sec:grTFT}, one component of our graph-based TFT is a vanilla \gls{tft} function (\ref{sec:bg_tft}) with four parameters itself. We evaluate three kinds of \gls{tft} defined here with notations of section \ref{sec:bg_tft}:
\begin{itemize}
    \item $TFT\alpha$: with just inertia $\alpha$ and a constant incentive $r_0$ : $(\alpha, \beta, \gamma, r_0) = (0.6, 0,0,0.3)$
    \item $TFT\beta$: addition of $\beta$ which adapts the incentive $r_t$ : $(\alpha, \beta, \gamma, r_0) = (0.6, 0.6,0,0.3)$
    \item $TFT\gamma$:  addition of $\gamma$ which provides stochasticity on $r_t$: $(\alpha, \beta, \gamma, r_0) = (0.6, 0.6,0.1,0.3)$
\end{itemize}
 To focus on the study of \gls{tft}, we run the simulations in a configuration where cooperation graphs are artificially communicated without detection (except for the last line which will be discussed in section \ref{sec:impact_detection}). We present the results of the social metrics in Table \ref{tab:results2}.

\begin{table}[h!]
\small
  \caption{Evaluation of parameters of grTFT (3 first lines) and the impact of cooperation detection (last line) in environments \textsc{Bilateral} and \textsc{Circular} (metrics $\times 100$, $3$ runs of $500$ steps ) } 
  \label{tab:results2}
  \centering
  
  \begin{tabular}{lcccccc}\toprule
  
& \multicolumn{3}{c}{\textsc{Bilateral}} & \multicolumn{3}{c}{\textsc{Circular}} 
\\
\cmidrule(lr){2-4}\cmidrule(lr){5-7}
           & $U$  & $IC$ & $Sf$    & $U$  & $IC$ & $Sf$ \\
           \midrule
$TFT\alpha$   & $\bm{66.7\pm 1.3}$ & $2.1\pm 4.0$ & $-6.6\pm 3.1$ & $64.5\pm 1.7$ & $2.9\pm 4.4$ & $-6.9\pm 2.9$\\
$TFT\beta$ & $62.7\pm 1.3$ & $\bm{14.7\pm 2.1}$ & $\bm{0.6\pm 1.7}$ & $61.3\pm 3.1$ & $13.3\pm 6.5$ & $\bm{1.3\pm 0.7}$ \\
$TFT\gamma$ & $62.8\pm 3.5$ & $14.5\pm 2.0$ & $-0.4\pm 0.5$ & $\bm{64.9\pm 3.2}$ & $\bm{14.5\pm 1.9}$ & $0.3\pm 0.6$  \\
$TFT\gamma_{-DET}$   & $59.9\pm 4.6$ & $-7.7\pm 2.9$ & $-13.2\pm 1.4$ & $57.9\pm 2.0$ & $-16.0\pm 5.7$ & $-14.1\pm 1.9$ \\\bottomrule
\end{tabular}
\end{table}

We can observe that the adaptive parameter $\beta$ is relevant to increase safety and incentive-compatibility, but it results in a slight drop of the efficiency $U$. This is solved by the stochastic parameter $\gamma$ which provides some forgiveness in order to reach a higher efficiency without decreasing safety.

\newpage

\begin{wrapfigure}[17]{r}{0.41\columnwidth}


        \includegraphics[trim=0.0cm 0.0cm 0.0cm 0.0cm, clip,width=0.40\columnwidth]{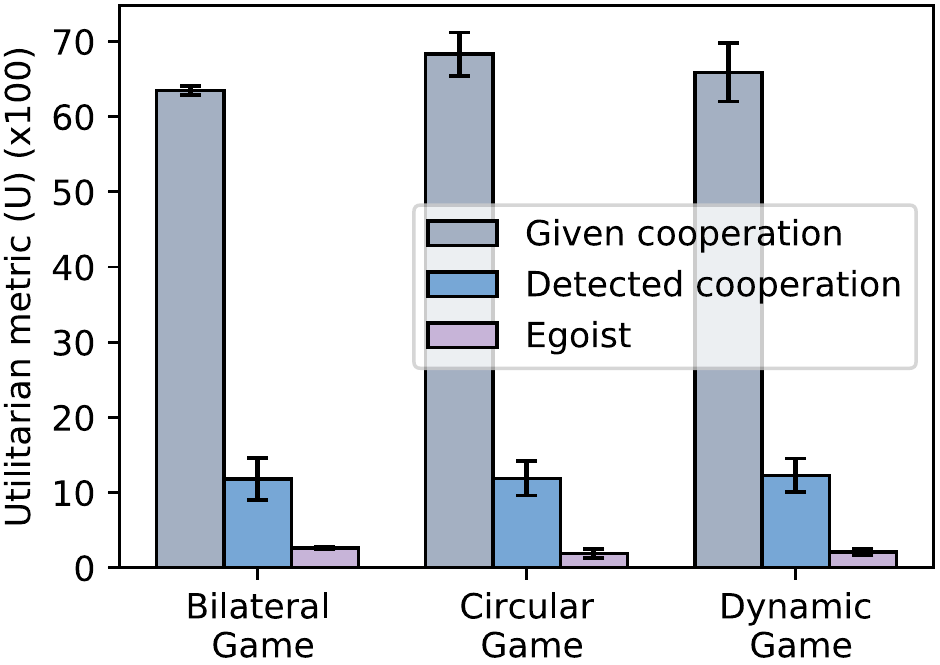}

 \caption{Impact of potential cooperation detection on the efficiency. $3$ runs of $500$ steps on three games: Bilateral, Circular and Dynamic (switch from a bilateral structure to circular one at $t=250$) }
  \label{fig:potential_impact}

\end{wrapfigure}

\subsection{Impact of cooperation detection}
\label{sec:impact_detection}

To evaluate the impact of cooperation detection, we compare two situations where the cooperation graphs (of maximal potential cooperation and of current cooperation) are artificially communicated or detected. For the detection of the current cooperation, a comparison is shown in the last two lines of Table \ref{tab:results2} ($TFT\gamma$ and $TFT\gamma_{\text{-DET}}$). The task is rather challenging and we can observe that without given cooperation, the agent struggles to be safe and incentive. However, regarding the detection of maximal potential cooperation (we recall that in the game \textsc{Collect}, it can be understood as the stochastic matrix which gives coin appearance probabilities), our approach manages to compute a reasonable graph\footnote{Simulation with detection of potential cooperation graph: \url{https://youtu.be/WhKLhflqZss}}. In Figure \ref{fig:potential_impact}, we show the utilitarian metric $U$ on some kinds of cooperation structure (fixed or dynamic). Indeed, although it is far from the oracle (when graph is given), our agent manages to reach cooperation since it performs better that mutual defection.



\subsection{Limitations}
\label{sec:limitations}
Since the agents have to detect the cooperative behavior of other agents as well as the potential cooperation graph, our approach, which currently struggles to properly detect cooperation, requires the full observation of the state and actions, which can be a major constraint in some situations.

Another limitation is the generation of \gls{rl} selfplay policies. In practice, we need to compute several policies with discrete cooperation degrees. Then, the computational cost of this offline training increases exponentially with the number of agents due to the combinatorial explosion. However, it could be interesting to study as further work the training of a unique policy per agent in which the continuous cooperation degrees vector would be embedded in the state.

\section{Conclusion}
\label{sec:conclusion}
In this paper we tackled the issue of sequential social dilemmas in situations where cooperation between agents are not necessarily bilateral, in particular when cooperation can be achieved with more than two players through a circular path. Since the related work in social dilemmas can't address this specificity, we proposed two main contributions:

\begin{itemize}
    \item The introduction and formalism of the circular sequential social dilemma: an extension of the Markov game introduced by \cite{leibo2017multi} allowing more complex cooperative relations between agents, including non-bilateral circular situations. As an example for experiments, we also provided the implementation of a simple game involving such dilemma and whose maximal potential cooperation graph is easily adaptable.
    \item A first approach to solve this kind of games: an agent using \gls{rl} selfplay policies of certain cooperation degrees governed by a graph-based \gls{tft}. This is an adaptation of the approach of \cite{lerer2017maintaining} which can address asymmetric and circular situations.
\end{itemize}

Even though our approach still struggles to optimally detect the cooperative behavior of agents, our results do demonstrate, however, the benefit of adding a graph-structure to \gls{tft} to address \gls{cssd} since vanilla \gls{tft} can only tackle symmetrical situations. 

We are excited to see future improvements and studies from the research community on such general-sum non-cooperative games. In particular, in view of the rising societal and environmental stakes, and the constant expansion of intelligent devices interacting with each other, actors seeking for competitiveness should strive to integrate collaboration in their environmental strategy to achieve better outcome and utility by optimally sharing existing resources. 

\newpage

\bibliographystyle{unsrt}
\bibliography{biblio}


\appendix

\section{Details of circular games}
\label{sec:experiments}
To study the \gls{cssd}  model and evaluate some parameters of our \textsc{grTFTrl} algorithm, we conducted some experiments on a multi-agent game. We designed this game in such a way that one can modify the graph structure of maximal potential of cooperation between the agents, in particular to confer the game a circular social dilemma structure (i.e. with non trivial cycles in the maximal graph of cooperation). In this section, we provide some details about the \textsc{Collect} game in Section \ref{sec:details} and then explain how to easily configure a maximal potential cooperation graph in Section \ref{sec:conf_graph}.

\subsection{Characteristics of the collect game}
\label{sec:details}
We briefly detail the \textsc{Collect} game. We recall that this game is a 4-player grid-world game in which coins appear and disappear stochastically. There are four rooms of size $W \times H$ in which one agent can move and collect either its own coins or those of another agent. At each time step, there is a fixed number of coins per room. The coins are either collected or stochastically disappear: each coin in the grid during $\tau$ steps disappears with probability $p(\tau) = [\tau - \tau_0]^+F(\tau - \tau_0)$ where $\tau_0 = W + H - 2$ and $F$ is the cumulative distribution function of the exponential law $Exp(\lambda = 1)$. If one coin is collected or disappears, another one appears randomly in the room and the selection of its color is given by a stochastic matrix (see in section \ref{sec:conf_graph}). There are five actions: right, down, left, up and still. The observations are binary tensors of size $(W, H, 2\times 4)$ where the layers indicate the location of the agents (the first 4 layers) and coins (the last 4 layers).

\subsection{Configurable cooperation structure}
\label{sec:conf_graph}
\begin{figure*}[h!]
\centering
  \captionsetup{justification=centering}
  
  \begin{subfigure}[b]{0.22\textwidth}
        \includegraphics[trim=0cm 7.5cm 22.5cm 0cm, clip,width=0.99\textwidth]{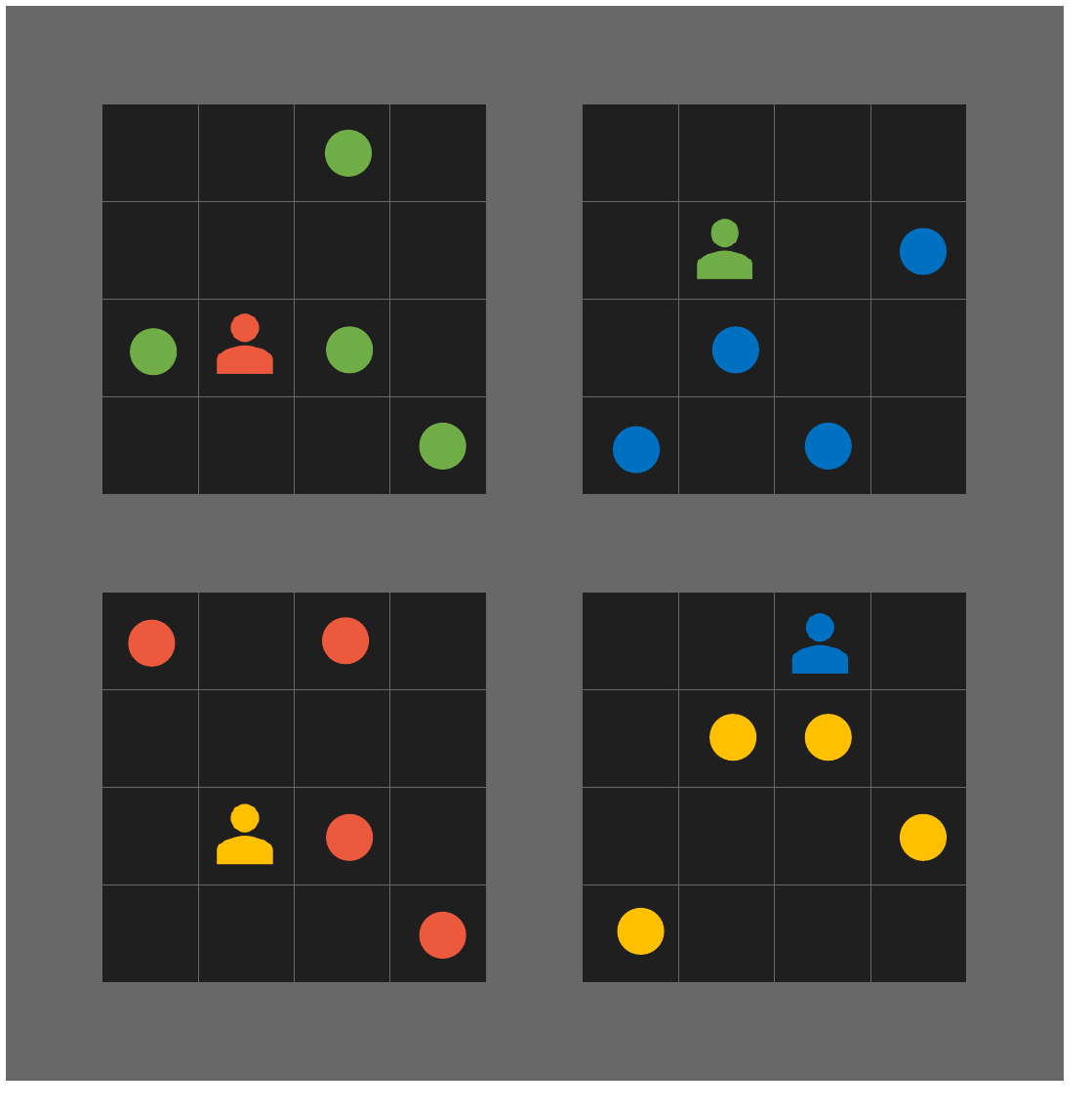}
 \caption{Circular}
  \label{fig:fetch_env_C}
  \end{subfigure}
~
 \begin{subfigure}[b]{0.22\textwidth}
        \includegraphics[trim=0cm 7.5cm 22.5cm 0cm, clip,width=0.99\textwidth]{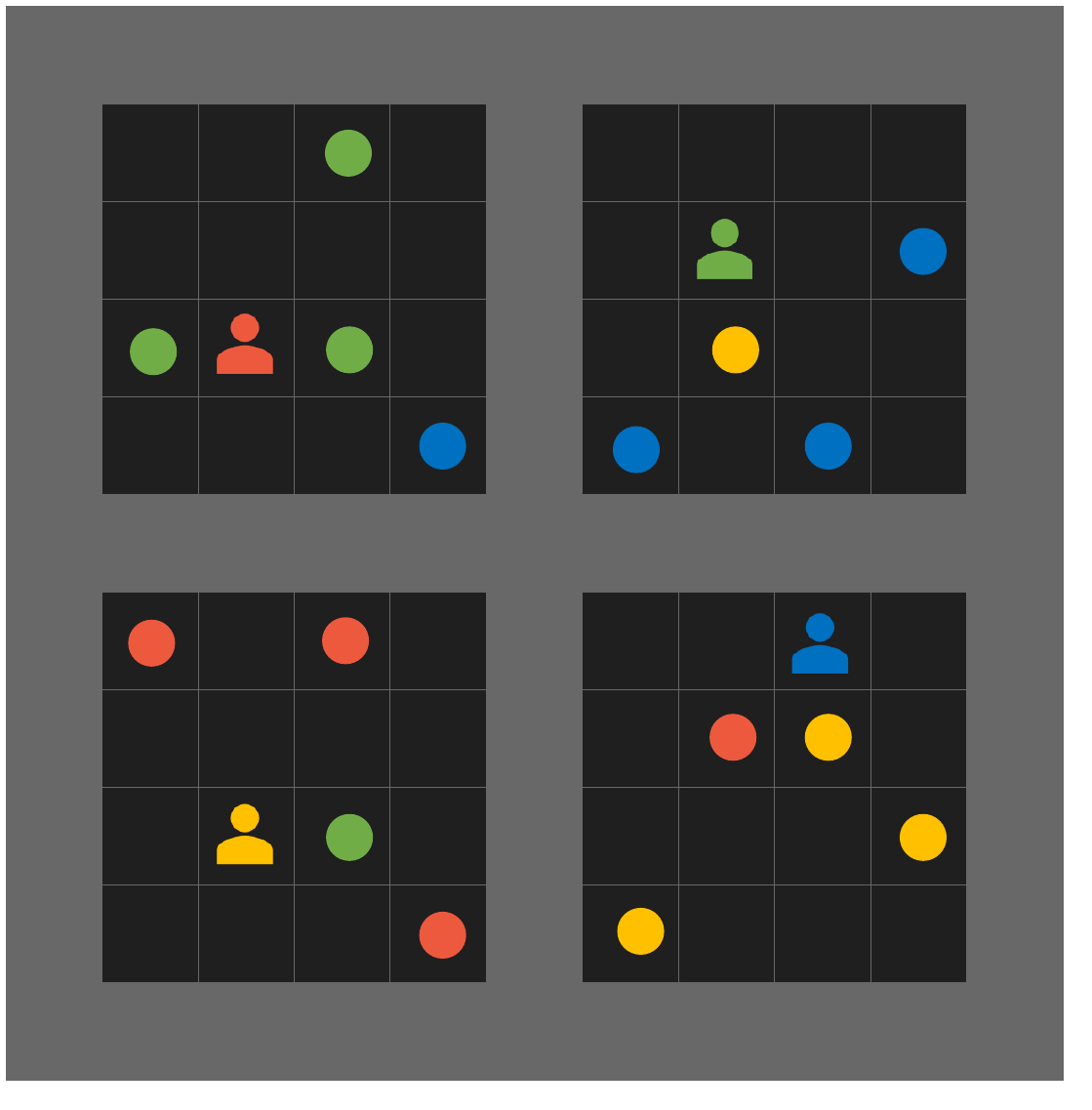}
 \caption{Semi-circular}
  \label{fig:fetch_env_S}
  \end{subfigure}
~
 \begin{subfigure}[b]{0.22\textwidth}
        \includegraphics[trim=0cm 7.5cm 22.5cm 0cm, clip,width=0.99\textwidth]{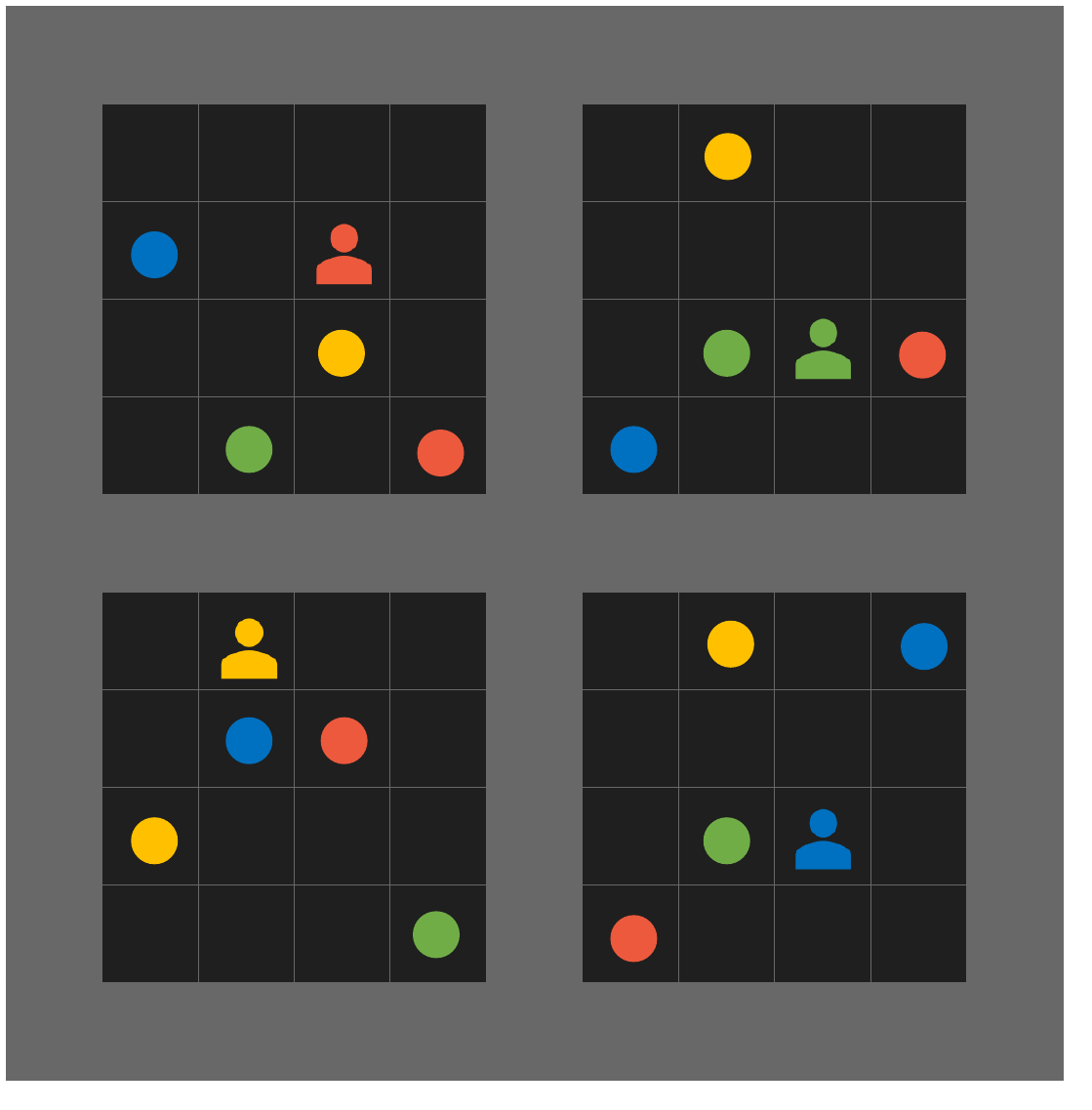}
 \caption{Full}
  \label{fig:fetch_env_F}
  \end{subfigure}
~
 \begin{subfigure}[b]{0.22\textwidth}
        \includegraphics[trim=0cm 7.5cm 22.5cm 0cm, clip,width=0.99\textwidth]{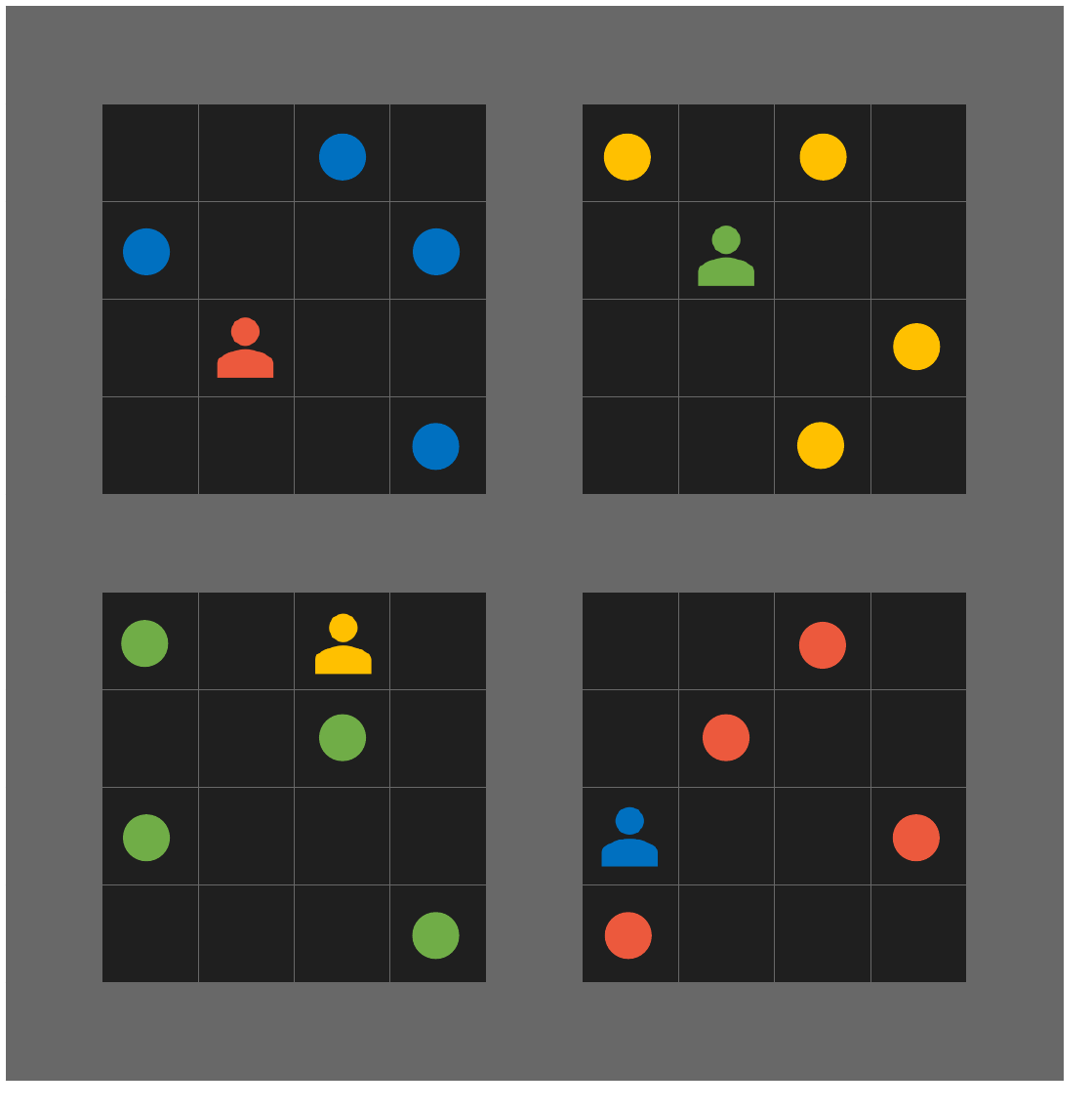}
 \caption{Bilateral}
  \label{fig:fetch_env_B}
  \end{subfigure}
  
  ~
  
 \begin{subfigure}[b]{0.22\textwidth}
 \centering
        \includegraphics[trim=10.8cm 13cm 17.6cm 0.1cm, clip,width=0.75\textwidth]{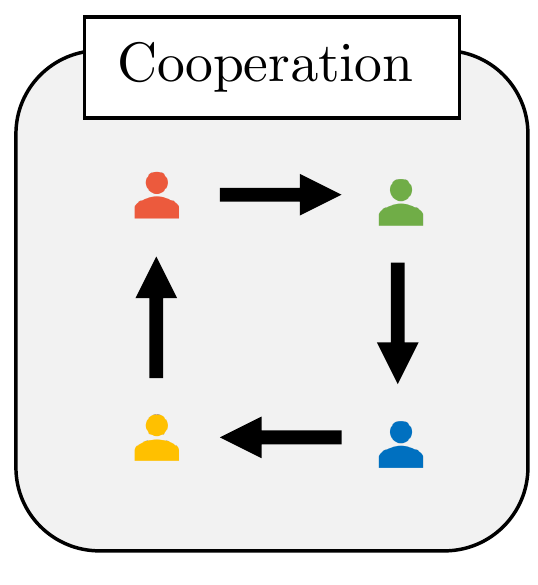}
 \caption{Cooperation graph \\ Circular case}
  \label{fig:fetch_gr_C}
  \end{subfigure}
~
 \begin{subfigure}[b]{0.22\textwidth}
 \centering
        \includegraphics[trim=10.8cm 13cm 17.6cm 0.1cm, clip,width=0.75\textwidth]{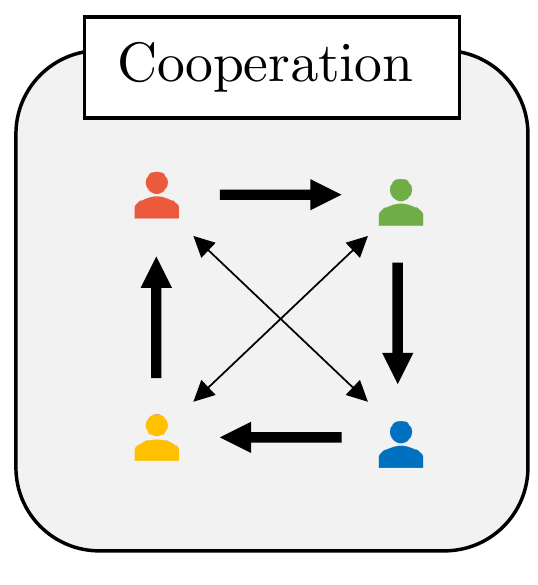}
 \caption{Cooperation graph \\ Semi-circular case}
  \label{fig:fetch_gr_S}
  \end{subfigure}
~
 \begin{subfigure}[b]{0.22\textwidth}
 \centering
        \includegraphics[trim=10.8cm 13cm 17.6cm 0.1cm, clip,width=0.75\textwidth]{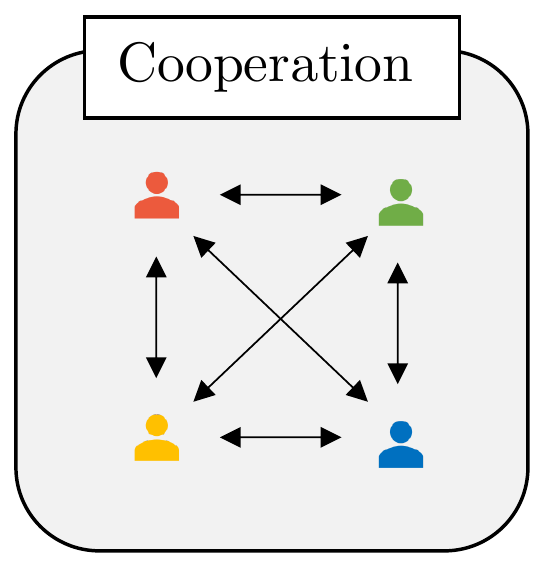}
 \caption{Cooperation graph \\ Full case}
  \label{fig:fetch_gr_F}
  \end{subfigure}
~
 \begin{subfigure}[b]{0.22\textwidth}
 \centering
        \includegraphics[trim=10.8cm 13cm 17.6cm 0.1cm, clip,width=0.75\textwidth]{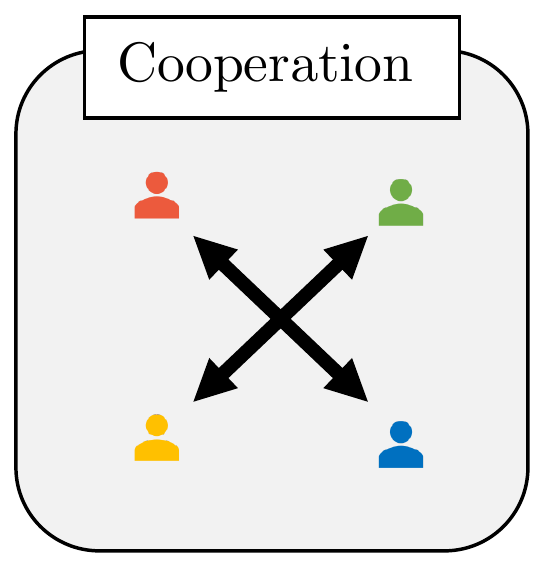}
 \caption{Cooperation graph \\ Bilateral case}
  \label{fig:fetch_gr_B}
  \end{subfigure}

 \caption{Game \textsc{Collect} with different structure of cooperation. There are four coins per room to better represent the stochasticity}
 \label{fig:Fetch_coop_graphs}
\end{figure*}

A key property of the \textsc{Collect} game is the configurability of the maximal potential of cooperation. Indeed, the probability of appearance of a coin of color $k$ in player $i$'s room is equal to $P[i,k]$ given by the matrix $P$ (which can be confounded with a weighted directed graph). Then, one can configure (or dynamically modify) $P$ to confer the game the wished cooperation structure. For example, the following matrices $C$, $S$, $F$ and $B$ correspond to the situations of Figure \ref{fig:Fetch_coop_graphs}. 
Matrix $C$ corresponds to an environment with a perfect circular cooperation (Figures \ref{fig:fetch_env_C} and \ref{fig:fetch_gr_C}); matrix $S$ is a semi-circular situation, reproducing the previous case with an alternative cycle  (Figures \ref{fig:fetch_env_S} and \ref{fig:fetch_gr_S}); matrix $F$ shows the full situation, a homogeneous case where each agent can cooperate with each other by pair in a equivalent way (Figures \ref{fig:fetch_env_F} and \ref{fig:fetch_gr_F}). At last, matrix $B$ represents a bilateral case where there are two distinct classic independent 2-player Sequential Social Dilemmas (Figures \ref{fig:fetch_env_B} and \ref{fig:fetch_gr_B}).

\[
C = \begin{bmatrix}
0 & 1 & 0 & 0 \\
0 & 0 & 1 & 0 \\
0 & 0 & 0 & 1 \\
1 & 0 & 0 & 0 
\end{bmatrix} \quad
S = \begin{bmatrix}
0.25 & 0.5 & 0.25 & 0 \\
0 & 0.25 & 0.5 & 0.25 \\
0.25 & 0 & 0.25 & 0.5 \\
0.5 & 0.25 & 0 & 0.25 
\end{bmatrix}
\] \newline
\[
F = \begin{bmatrix}
0.25 & 0.25 & 0.25 & 0.25 \\
0.25 & 0.25 & 0.25 & 0.25 \\
0.25 & 0.25 & 0.25 & 0.25 \\
0.25 & 0.25 & 0.25 & 0.25
\end{bmatrix} \quad
B = \begin{bmatrix}
0 & 0 & 1 & 0 \\
0 & 0 & 0 & 1 \\
1 & 0 & 0 & 0 \\
0 & 1 & 0 & 0 
\end{bmatrix}
\]

\newpage
\section{Graph-based Tit-for-Tat}

The algorithm introduced in this paper needs a graph-based Tit-for-Tat (TFT) component, an extension of the Tit-for-Tat \cite{rapoport1965prisoner, verhoeff1998trader} with a graph structure whose goal is to react to the various cooperation degrees between agents. As we designed this algorithm in a previous submission still in review, we summarize it this section.

\subsection{Main principle}
Let's consider $N$ agents in a social dilemma whose maximal potential cooperation is given by a weighted directed graph $\mathcal{G}_{max} \in [0,1]^{N \times N} $. The goal of the graph-based Tit-for-Tat is that at each step $t$, an agent $k$ reacts to the previous cooperation degrees $\mathcal{C}^{t-1} \in [0,1]^{N \times N}$ and computes a vector $\overrightarrow{C_k} \in [0,1]^{N}$ in reaction of the behavior of the other agents.
\begin{equation}
     \overrightarrow{C_k} = \text{grTFT}(t, k, \mathcal{G}_{max}, {\mathcal{C}}^{t-1})
\end{equation}

\subsection{Components}

A grTFT agent constantly updates two components:

\begin{itemize}
    \item A max source flow $\mathcal{D}_k$ indicating the maximal effort of cooperation an agent wants to offer.
    \item An inner cooperation graph $\mathcal{C}_k$ in which the agent searches the maximal flow of cooperation given by $\mathcal{D}_k$.
\end{itemize}

The agent's goal is to adapt $\mathcal{C}_k$  to estimate the directed cooperation relations $(i,j)$ as well to measure how to incentive cooperation or penalize defectors. With the TFT function, the agent also adapts its maximal effort of cooperation $\mathcal{D}_k$ according to the global cooperation dynamics. Once those updates have been performed, the principle if to find the optimal cycle through $\mathcal{C}_k$ with a maximal capacity given by $\mathcal{D}_k$ (in particular trivial bilateral 2-players cycles).

Note that in the case of our paper, the agent's maximal potential graph can be regularly updated. However, without loss of generality, we fix it constant over time in the following sections.

\subsection{Parameters}
A grTFT agent has two main parameters:

\begin{itemize}
    \item A TFT function $f_{TFT}: [0,1]\times [0,1] \rightarrow [0,1]$ such that $a^{t+1} = f_{TFT}(a^{t}, b^{t})$ is the ideal response according to a detected degree $b^{t}$ and the previous response $a^{t}$
    \item A flow network algorithm: to find the cycle of maximal flow (cooperation). For example, Ford-Fulkerson \cite{ford1956maximal} (shortest path) or with constraints \cite{orlin1997polynomial} (longest path: more pro-social without loss of utility)
\end{itemize}

\subsection{The grTFT algorithm}

The grTFT algorithm is divided into several phases \cite{anonym2021tft}. At each step:

\begin{enumerate}
    \item For each other player $j$, update $\mathcal{C}_k[k,j]$ with $f_{TFT}$ according to the difference between what $j$ "received" and what he "gave" at previous step.
    \item Update the source flow $\mathcal{D}_k$ with $f_{TFT}$ according to the difference between what $k$ (oneself) "has received" and what $k$ (oneself) "had given" at previous step.
    \item Create a flow network $\mathcal{F}$, whose capacities are given by $\mathcal{C}_k$ with a source vertex directed towards the vertex of $k$ (oneself) with capacity $\mathcal{D}_k$ and all edges of $\mathcal{C}_k$ initially directed towards vertex $k$ artificially redirected towards a sink vertex. Thus, we have a flow network allowing to find the maximum cyclic flow (i.e. from $k$ to $k$). 
    \item Compute the maximum flow $\mathcal{R}$ on $\mathcal{F}$, i.e. a sub-graph in $\mathcal{F}$ and extract the next choice of cooperation $\overrightarrow{C_k} \leftarrow \mathcal{R}[k,:]$
\end{enumerate}

\begin{algorithm}[H]
\caption{\textsc{grTFT} (for agent $k$)}
\SetAlgoLined
\textbf{Input:} Max cooperation graph $\mathcal{C}_{max}$ and max source flow $D_{max}$ given by the game and a TFT function $f_{TFT}$ \\
Initialize: $\mathcal{C}_k \leftarrow \mathcal{C}_{max}, ~ \mathcal{D}_k \leftarrow D_{max}[k] $\\
First step: Choose $\forall j \neq k, ~ \Vec{C_k}[j] \leftarrow f_{TFT}(t=0)$\\
 \For{$t \in [1, T_{max}]$}{
 \For{each other agent $j$}{
 From $C^{t-1}$, compute outgoing flow of cooperation of $j$ : $(C_{j}^{t-1})^+$ \\
 Execute a TFT on $j$ : $c_{kj}^{t} = f_{TFT}(c_{kj}^{t-1}, (C_{j}^{t-1})^+) $ \\
 Modify the inner cooperation graph: $\mathcal{C}_k[k,j] \leftarrow c_{kj}^{t}\mathcal{C}_{max}$ \\
}
From $C^{t-1}$, compute the incoming flow of cooperation for $k: (C_{k}^{t-1})^-$\\
Update by TFT the next source flow: $\mathcal{D}_k \leftarrow f_{TFT}(\mathcal{D}_k, (C_{k}^{t-1})^-) $\\
Generate a new flow network $\mathcal{F}$ from $k$ to $k$ with a source of capacity $\mathcal{D}_k$ and capacities given by $\mathcal{C}_k$  \\
From $\mathcal{F}$, extract the sub-graph $\mathcal{R} $ of maximum flow of cooperation  \\
Choose cooperation degrees from max flow: $\overrightarrow{C_k^{t}} \leftarrow \mathcal{R}[k,:] $}
\label{algo:grTFT}
\end{algorithm}

\section{Further implementation details}

\subsection{Some parameters of our \textsc{grTFTrl} agent}
In this section, we give some parameters used in our simulations. For stability, discrete cooperative policies are selected every $K = 5$ time steps. In further work, it could be a parameter to evaluate. The soft-updates of detection functions use a coefficient $\tau = 0.1$.

\subsection{DQN training}
The choice of the best \gls{rl} algorithm was out of scope of the paper. We used a simple DQN with two layers of 150 and 100 neural units, and the state was restrained to the room of a agent. For computational economy, all agents used the same cooperative policy with a preprossessing function permuting the layers to represent the view of one agent cooperating with another agent. 

\subsection{Computational aspects}
The main goal of the paper was to design a hybrid algorithm with \gls{rl} policies and \gls{tft} strategies, to evaluate some functions, parameters and explore the possibilities of such model. Therefore, we insisted on the possibility to train and execute our simulations without significant computational resources. The instances of the games were small grid-world making the training possible with a simple laptop. The graphical rendering of execution phases can require several seconds per timestep, but it can be disabled.

\section{Code and videos}

\subsection{Implementation of our \textsc{grTFTrl}}

The implementation of our agent is available here in Python:\newline
\url{https://github.com/submission-conf/neurips_cooperativeAI}\newline

It is possible to select some preset simulations with different configurations of the game (\textsc{Circular}/\textsc{Bilateral}) and different types of agents (grTFT, Vanilla TFT, Egoist, Nice). An evaluation script with the implementation of our social metrics (efficiency, incentive-compatibility and safety) is also available.

The game is based on a multi-agent version \cite{gym_multigrid} of a minigrid environment \cite{gym_minigrid}.

\subsection{Simulations videos}
Some videos of a few simulations are available here:\newline \url{https://youtube.com/playlist?list=PLzmQsQrITrGI16_-nj2qSzgTMHN1hCIoq} \newline

The videos show the agents' behavior during the game, as well as the evolution of the key elements: the sum of payoffs, the maximal cooperation graph, the detected potential graph (useful if not given), the current choice of cooperation between agents (extracted from the different agents \textsc{grTFTrl}) and the detected current cooperation graph (if not given).

\end{document}

%% file: acronyms.tex
\newacronym{cssd}{CSSD}{Circular Sequential Social Dilemma}
\newacronym{ssd}{SSD}{Sequential Social Dilemma}
\newacronym{pd}{PD}{Prisoner's Dilemma}
\newacronym{ipd}{IPD}{Iterated Prisoner's Dilemma}
\newacronym{gipd}{GIPD}{Graph-based Iterated Prisoner's Dilemma}
\newacronym{tft}{TFT}{Tit-for-Tat}
\newacronym{gtft}{GTFT}{Graph-based Tit-for-Tat}
\newacronym{marl}{MARL}{Multi-Agent Reinforcement Learning}
\newacronym{rl}{RL}{Reinforcement Learning}